# Vietnamese Legal Ontology: Every legal answer, traced to the article that grounds it

KHAC-HOAI NAM BUI and QUAN TRAN MINH

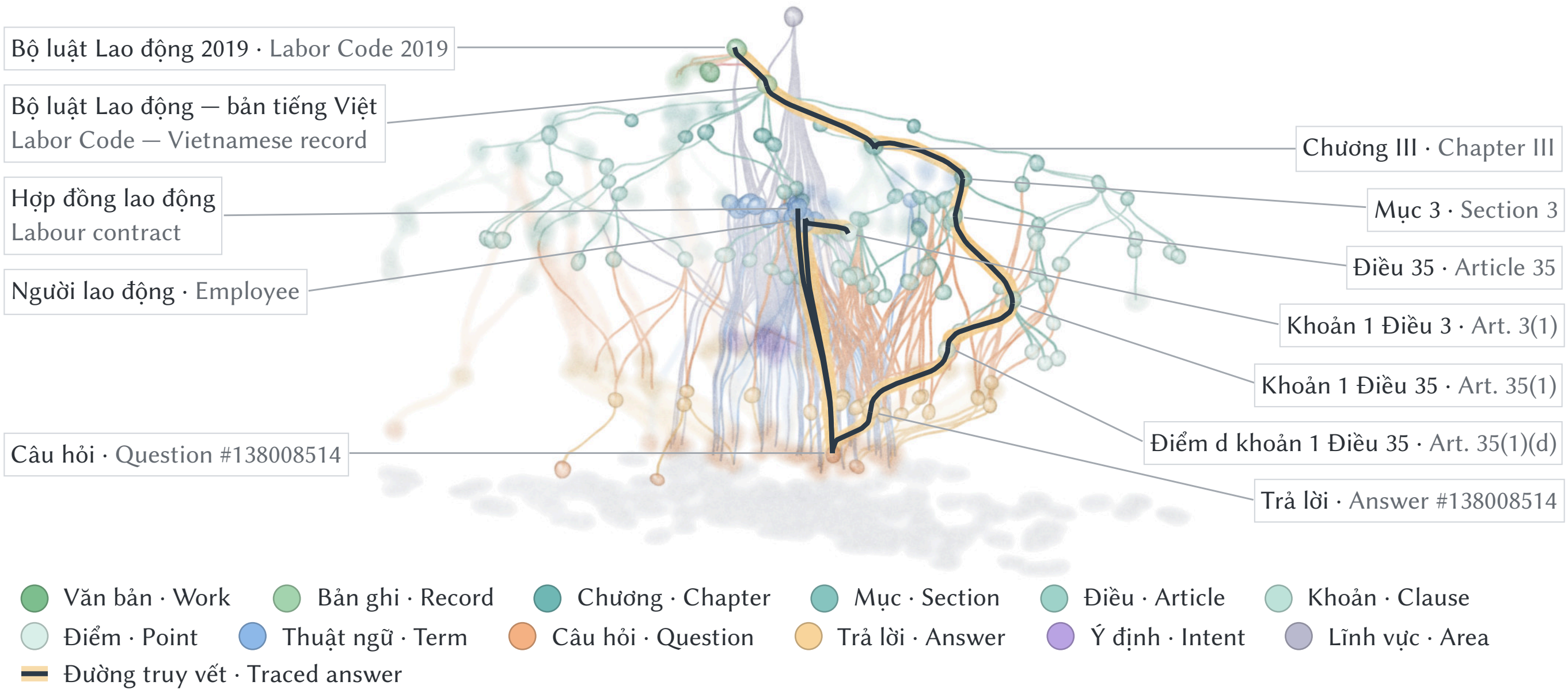

Fig. 1. **One answer, traced through the graph.** Q&A #138008514 asks how much notice an employee must give before resigning in 2024. Around it, the graph holds 229 nodes and 454 edges of 12 classes: the answer and the 7 sources it cites, the parsed chapters, articles, clauses and points of the 2019 Labor Code, the terms the question mentions, and 43 other questions that cite the same code. The dark path is the only citation the answer quotes word for word: point d of Article 35(1).

A legal answer is only as good as the article it rests on. Vietnam's largest open legal corpus pairs 109,543 published questions and answers with 573,851 legal documents, yet its citations are free text: *"khoản 3 Điều 101 của Bộ luật này"* ("Clause 3, Article 101 of this Code") names neither the code nor its version. Following the five stages of NVIDIA Auto Ontology, Discover, Ground, Validate, Expose and Learn, we turn four open datasets into a knowledge graph of 11.5 M nodes and 16.0 M edges in which every node and edge carries its source record and method, with no duplicate keys, dangling edges or schema violations. Of 67,868 cited provisions, 61,487 resolve to a dated law version, 51,694 to a parsed article, clause or point, and 19,012 are confirmed word for word. The ontology then shows what flat records hide. 36% of the 78,361 answers whose citations resolve now cite a law that is no longer in force; 2,525 verified citations point to an article number that says something else in the current law; 2,520 citations were re-versioned because their words match a different version than their date suggests; and 5,575 document numbers are shared by several local authorities, one of them by 72. Every number in this paper is computed from the graph.

## 1 Introduction

On 14 Mar 2024, a reader of Vietnam's largest legal portal asked: *"Cách tính thời gian báo trước khi nghỉ việc của người lao động năm 2024? Báo trước 30 hay 45 ngày?"* ("How to calculate the notice period for an employee's resignation in 2024? 30 or 45 days' notice?") The published answer cites 7 sources. Which of them grounds it? Only one is quoted word for word: point d of Article 35(1) of the 2019 Labor Code, Law 45/2019/QH14. Five are links to related articles whose text the answer does not use, and one is a decree cited as a whole. A reader cannot tell these apart, and neither can a retrieval system that indexes the answer as plain text [Lewis et al. 2020].

The difficulty is not this answer; it is how legal knowledge is linked. The four open datasets we study [tmquan 2026] hold 573,851 legal documents, 109,543 questions with answers, 17,091 legal terms and 681 legal-assistance articles, all drawn from `thuvienphapluat.vn`. Their links are strings. An answer writes *"Điều 35 Bộ luật Lao động"* ("Article 35, Labor Code") without saying which Labor Code, or *"Bộ luật này"* ("this Code") and leaves the antecedent to the reader. Many answers were written under laws that have since been replaced. One document number can name decisions of many different provinces, and one term can be defined in dozens of laws.

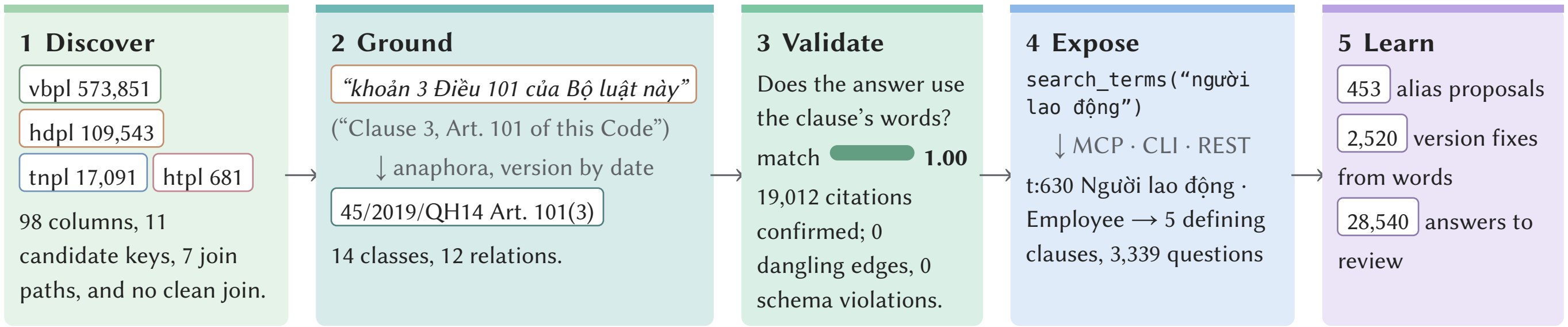

Fig. 2. **From strings to grounded edges, in five stages.** The running example: an answer's *"Bộ luật này"* ("this Code") becomes an edge to Clause 3 of Article 101 of the 2019 Labor Code, checked against the answer's words, served to agents, and every failure becomes a reviewable proposal.

We make these links explicit and checkable. Following the five stages of NVIDIA Auto Ontology [NVIDIA 2026], we *discover* the data, *ground* it in an ontology, *validate* every assertion, *expose* the result to people and agents, and *learn* from what fails (Fig. 2). The result is a knowledge graph [Hogan et al. 2021] in which every answer points to the exact article, clause or point it cites, in the version in force when the answer was published, together with the evidence for that link (Fig. 1). The graph has 11,526,901 nodes and 15,958,034 edges in 14 classes and 12 relations. Every node and edge records its source dataset, source record and method; none is duplicated, dangling or outside its relation's domain and range. It rebuilds from the raw datasets in about 4.3 minutes on one node.

The graph matters less than what its ontology lets us see. Once a citation is a typed edge to a dated version of a parsed article, questions that were out of reach become short queries. Which answers have aged since they were written? Which article numbers changed meaning between versions? Which version did a writer actually have in mind? We contribute:

- an ontology for Vietnamese law and legal Q&A that separates a law from its records, its structure from its metadata, and every claim from its evidence, aligned with ELI, Akoma Ntoso, FRBR, SKOS, schema.org and PROV-O (Section 3);
- a grounding and validation cascade that resolves 91% of 67,868 cited provisions to a dated law version, recovers 92% of the citations that name no law from their context, and confirms 19,012 citations by their words (Section 4);
- findings that only the graph can show: what the ontology reveals about identity, definitions and structure (Section 6), answers that age (Section 7.1), article numbers that change meaning (Section 7.2), and words that reveal which version a writer meant (Section 7.3);
- three ways to query the graph, for agents (MCP), the command line and the web (REST), and a learning loop that turns failures into reviewable proposals (Section 8).

## 2 Four datasets, one legal corpus

Table 1 lists the inputs. The legal documents carry full text with metadata: number, type, issuing authority, dates and a status line; 15,495 of them have an English record. Each Q&A record holds a question, an answer, and the law names and article strings the answer cites. Each term has a definition and a legal area. The legal-assistance records are articles, not dialogues: advice, explanations of disputes, and summaries of new policy.

Table 1. Four open datasets from `thuvienphapluat.vn`, published on Hugging Face under `tmquan/thuvienphapluat-vn-*`.

| Dataset | Name · English | Records |
|---|---|---|
| `vbpl` | *Văn bản pháp luật* · Legal documents | 573,851 |
| `hdpl` | *Hỏi đáp pháp luật* · Legal Q&A | 109,543 |
| `tnpl` | *Thuật ngữ pháp lý* · Legal terms | 17,091 |
| `htpl` | *Hỗ trợ pháp lý* · Legal assistance | 681 |

Discover profiled 98 columns and found 11 candidate keys and 7 join paths between the four datasets, and no join among them is clean. The Q&A names laws and articles in prose, not by identifier. Terms carry no link to the clause that defines them. Legal areas follow two schemes of different granularity (45 ↔ 34 areas). Grounding therefore starts by designing the targets that a citation can resolve to.

## 3 An ontology that answers can point into

Fig. 3 shows the ontology: 14 classes in three groups (the law and its structure, concepts, and usage) linked by 12 relations, each aligned with an established vocabulary (Section B). Four decisions shape it.

**A law is not a record.** A `LegalWork` is a law identified by its number; a `LegalDocument` is one record of it (the Vietnamese text, an English translation or a consolidated version) and `realizes` the work, as an expression realizes a work in FRBR [IFLA Study Group on the Functional Requirements for Bibliographic Records 1998] and in ELI [Council of the European Union 2012]. Reconciliation folds 573,851 records into 550,654 works, and 20,096 works merge two or more records (43,293 records in all). Numbers alone are not keys: 231,510 local documents become unique only once their number is qualified by the issuing authority, read from the metadata or from a header such as *"ỦY BAN NHÂN DÂN TỈNH …"* ("People's Committee of … Province"). A further 862 records whose type contradicts the code in their number are split off.

**Structure is data.** Every text is parsed into *Chương* › *Mục* › *Điều* › *Khoản* › *Điểm* (Chapter › Section › Article › Clause › Point), the hierarchy of Akoma Ntoso [Palmirani and Vitali 2011]. The graph holds 2,213,352 articles, 4,583,894 clauses and 3,073,777 points joined by 10,165,016 `contains` edges, so a citation can land on a point rather than on a whole law.

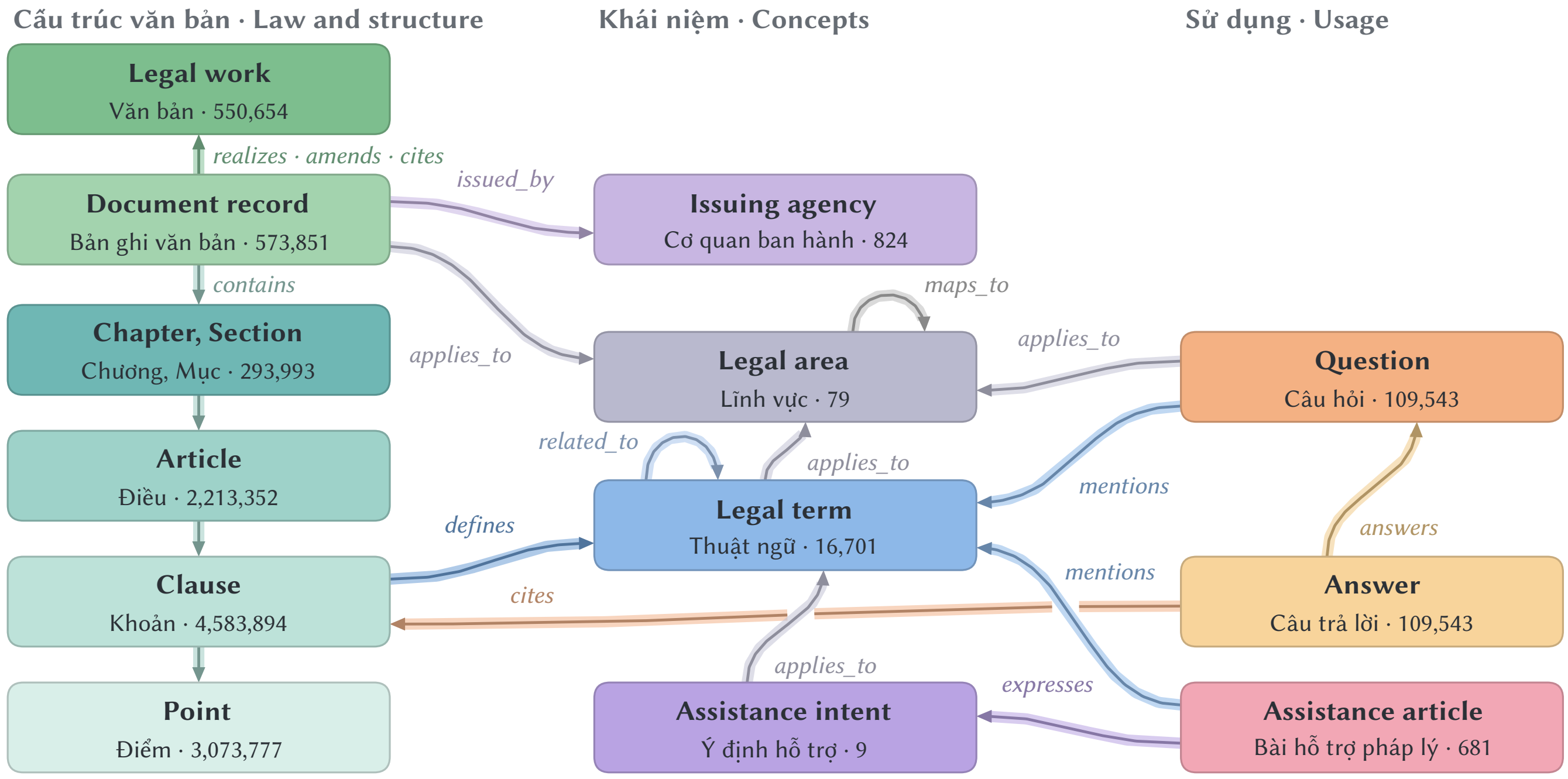

Fig. 3. **The ontology separates a law from its records, its structure from its metadata, and every claim from its evidence.** Fourteen classes in three groups and twelve relations; each box gives the class in English and Vietnamese with its node count in the graph, and each arrow runs from domain to range. Chapter and Section share a box.

**Time is a relation.** Laws of the same type and name form version families ordered by effective date (336 families, 102 with more than one version). Repeals are read from the text itself, in sentences where a law declares that another *"... hết hiệu lực / bãi bỏ / thay thế"* ("... expires / is repealed / is replaced"), and are kept only if they agree with rank and metadata: 30,618 whole repeals are kept and 1,061 rejected.

**Every claim carries its evidence.** Each edge records the method that produced it, its source record and a status, in the spirit of PROV-O [Lebo et al. 2013]. Of 15,958,034 edges, 15,612,341 are validated, 302,211 are candidates awaiting review, 33,689 are structural and 9,793 are flagged.

Knowledge classes tie the datasets together. Terms are SKOS concepts [Miles and Bechhofer 2009], linked to the clauses that define them by the definitional pattern *"X là ..."* ("X is ...") through 47,691 `defines` edges, and to the questions that mention them (242,149 `mentions` edges). The two area schemes are mapped with SKOS relations (SKOS: 22 exact · 7 close · 3 narrow · 3 related). Assistance articles are typed by 9 intents read from their category and title, for example *"tranh-chap"* (disputes) and *"chinh-sach-moi"* (new policy); questions and answers use schema.org types [Guha et al. 2016].

## 4 Grounding a citation, and proving it

A citation in an answer is a claim: this answer rests on that provision. Validation turns the claim into an edge only after answering three questions (which law, which version, which provision) and then asks a fourth: does the answer actually use the provision's words? Fig. 4 follows the 67,868 cited provisions through these steps; Section C lists every rule.

**Which law?** Only 19,219 citations name their law; 48,649 do not. Named ones resolve by exact name, unique prefix or fuzzy match. For the rest, the resolver reads the context the way a lawyer would: a law name next to the article number, an anaphora such as *"... này"* ("this ..."), the nearest preceding reference of the same type, the only law the answer cites, or the question itself. It recovers 44,926 of 48,649 (92%), 24,651 of them through anaphora. Answer #138017201 (10 Jan 2024) asks whether unused annual leave can be carried over and writes *"khoản 3 Điều 101 của Bộ luật này"* ("Clause 3, Article 101 of this Code"). The antecedent, *Bộ luật Lao động 2019* (Labor Code 2019), appears 145 characters earlier, so the citation resolves to Clause 3 of Article 101 of Law 45/2019/QH14, whose text the answer quotes in full (match 1.00).

**Which version?** An explicit number decides first, then an explicit year, then the version in force on the publication date, and otherwise the only version: 27,032, 199, 25,551 and 6,185 citations respectively. When another version of the same law matches the answer's words at least 0.5 and by a margin of 0.25, the words win; 2,520 citations were re-versioned this way (Section 7.3).

**Which provision?** The *Điều / Khoản / Điểm* (article / clause / point) path is looked up in the parsed hierarchy of the chosen version. 51,694 citations are found; 5,683 name an article absent from that version, and 4,110 a version with no parsed structure.

**Are the words there?** We measure how much of the provision's text appears in the answer as word 5-grams. 19,012 citations reach 0.5 and are validated, 28% of all cited provisions. The rest stay structural links: kept, labelled and queued, never silently promoted.

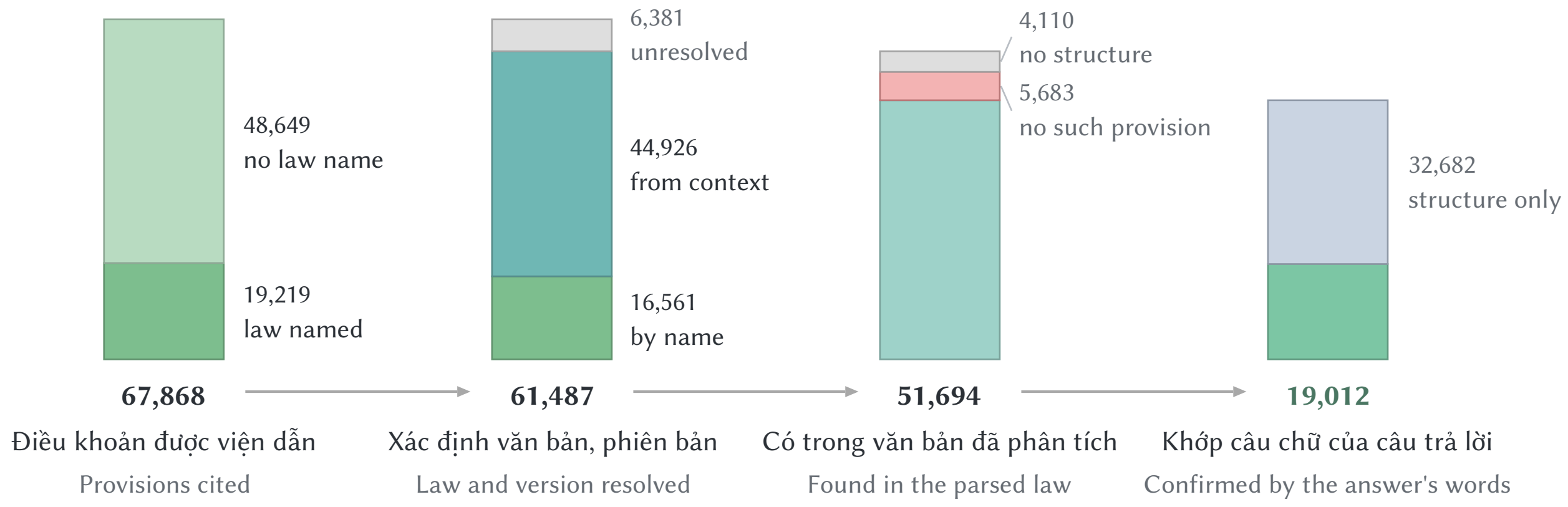

Fig. 4. **From 67,868 cited provisions to 19,012 confirmed by their words.** Each bar is what survives a step; hatched blocks are what drops out, and why. Most citations name no law at all; 92% of those are recovered from their context before a version is chosen.

**Is the graph sound?** Every check over the whole graph passes: 0 duplicate node keys, 0 dangling edges, 0 domain or range violations among 15,958,034 typed edges, and 0 nodes or edges without provenance.

Fig. 5 shows the result for the question in the introduction. On 14 Mar 2024 the 2019 Labor Code, not the 2012 one, was in force, so the citation resolves to Law 45/2019/QH14. Point d of Article 35(1), which leaves the notice period for special sectors to the Government, appears in the answer word for word, and the answer duly cites the Government's Decree 145/2020/NĐ-CP as well. The question's two key terms, *người lao động* (employee) and *hợp đồng lao động* (labour contract), are linked to the clauses that define them. The command `vnkg trace a:138008514` prints the same chain as 43 provenance hops (Section 8).

## 5 A gallery of verdicts

The trace of Fig. 1 is not a special case. Fig. 6 repeats it for twelve more answers, one for each of the laws in Fig. 12, published between 2017 and 2026.

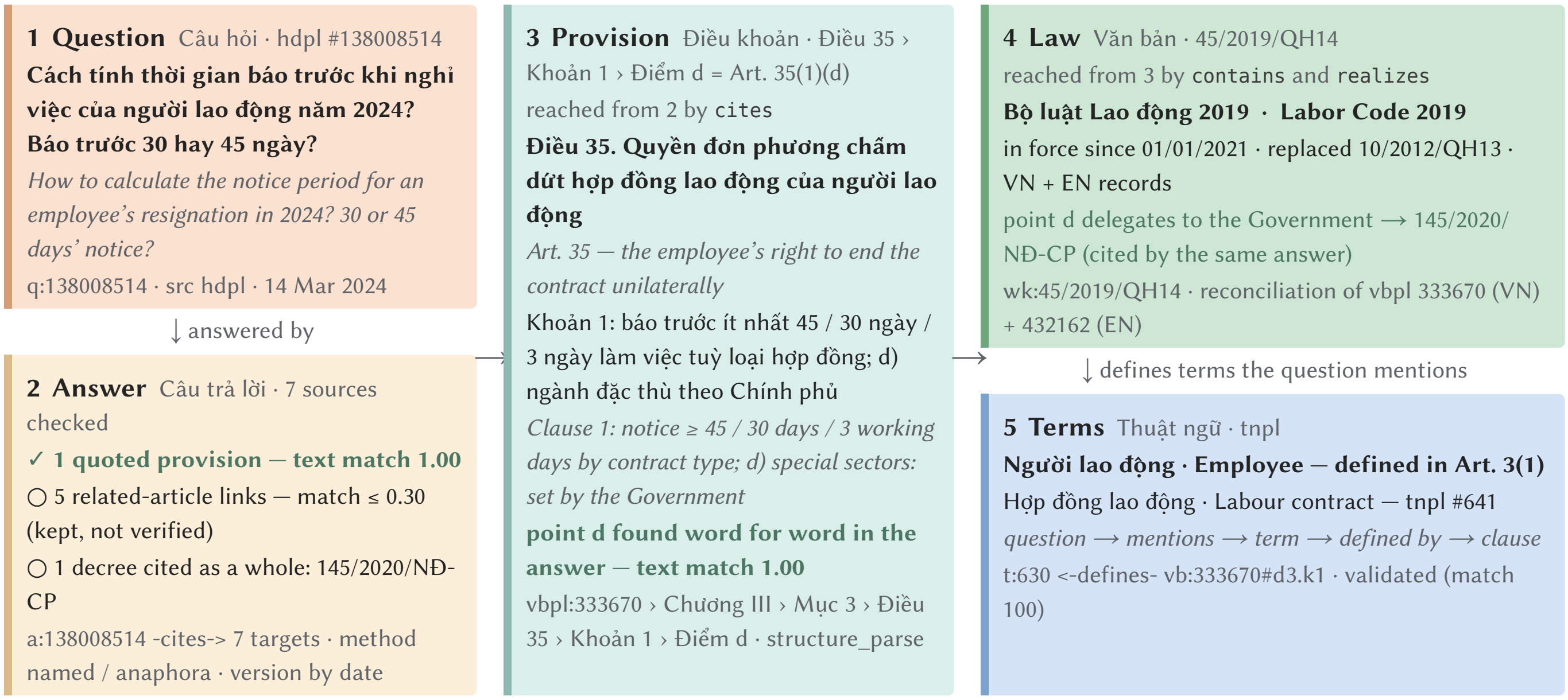

Fig. 5. **The trace behind Fig. 1, hop by hop.** Each box is a node with its provenance: source dataset and record, method and status. Of the 7 sources the answer cites, only point d of Article 35(1) appears in it word for word.

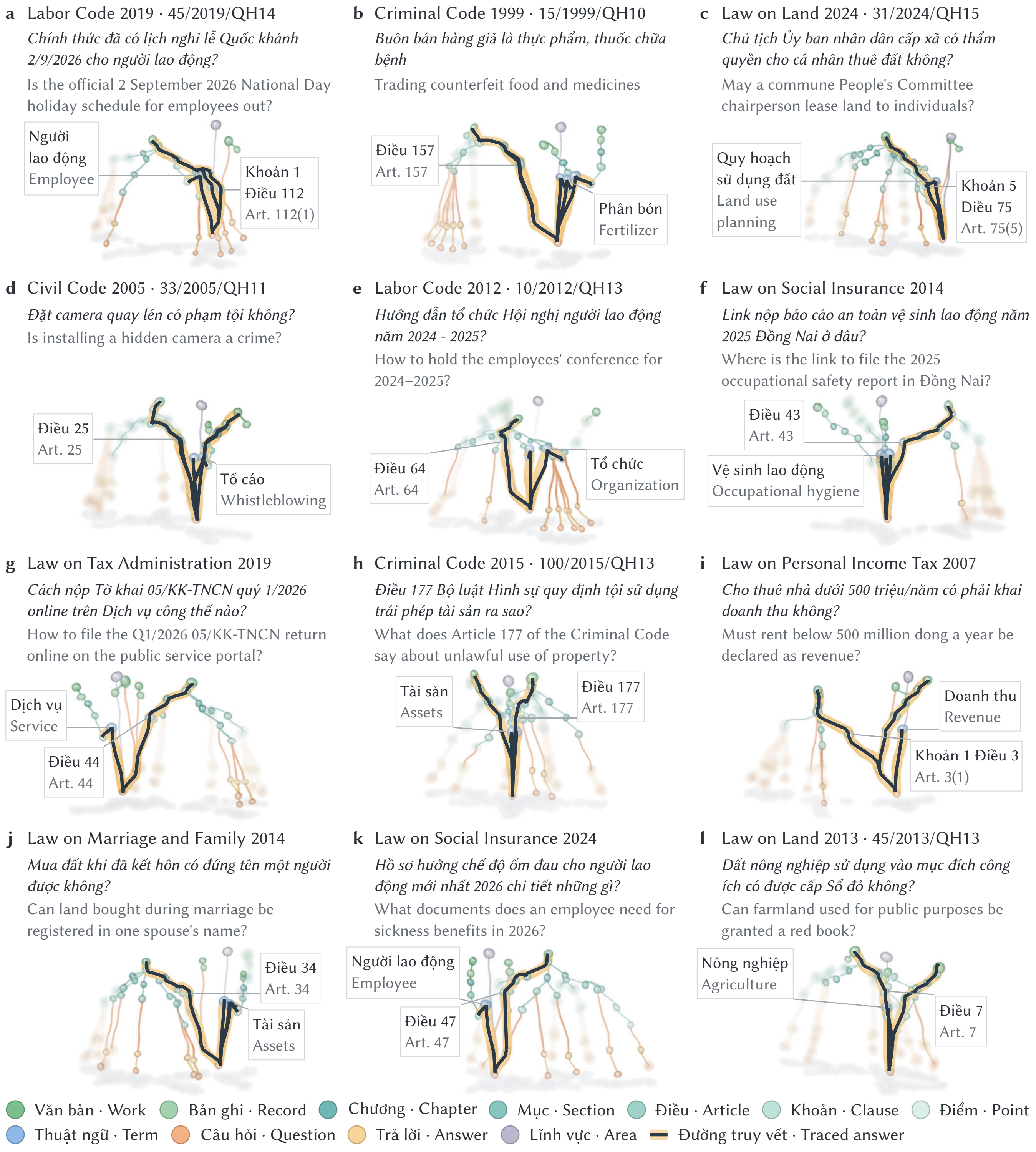

Fig. 6. **Twelve more answers, traced.** One answer for each of the twelve laws with the most text-verified answers, drawn like Fig. 1: question, sources, the chain up to the law, terms with their defining clauses, and other answers citing the same law (38 to 60 nodes per panel; numbers and dates in Table 10). Dark paths lead to the sources whose text the answer contains (match at least 0.5); labels name the provision it quotes most closely and one term.

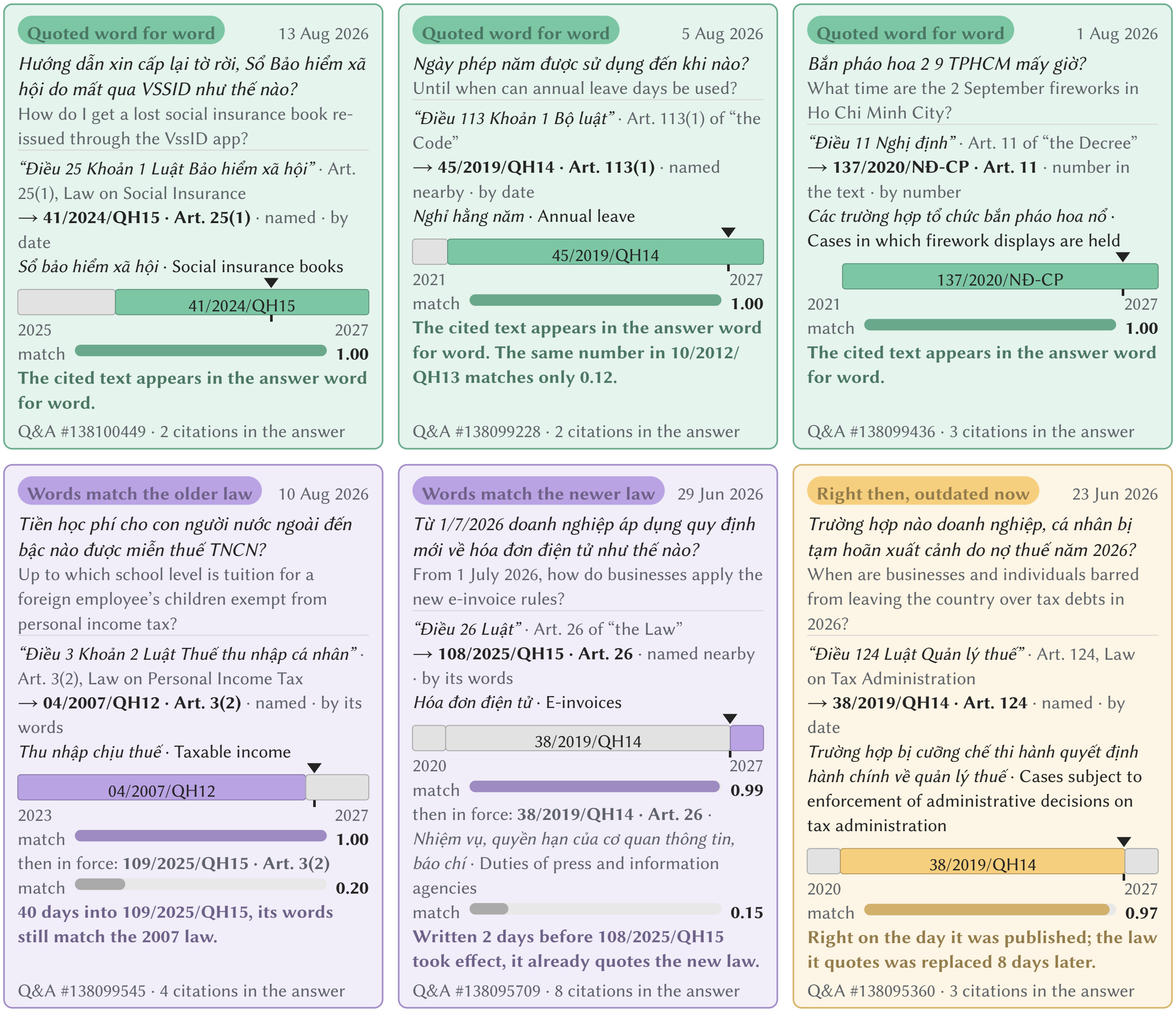

Fig. 7. **Six citations confirmed by their words.** Each card is one citation from a 2026 answer as the graph checks it: law, version and article; when each version was in force (bars; triangle: publication date); and how much of the provision's text the answer contains (grey: the version then in force, where the words chose another). Top: three ways of naming a law. Bottom: versions chosen by the words, both ways, and one replaced days later.

In two of them every source the answer cites is confirmed by its words; in the others the trace separates the provisions the answer quotes from the sources it only names.

Validation does not end in a yes or no. Fig. 7 and Fig. 8 show nine recent answers and the verdict the graph reaches for one citation in each. Three quote their provision word for word: one names its law, one says only *"Bộ luật"* ("the Code"), and one cites a decree by its number. Two are pinned to a version by their words rather than their date, one backwards and one forwards, around the day new tax laws took effect. One was right when published and outdated within days. The graph flags the other three (Fig. 8): one quotes a labor code repealed years before it was written, one paraphrases a clause that does exist, and one cannot be resolved because the law's name was extracted with a typo and a stray word. Each card is generated from the graph: the version bars come from the dataset's dates, the match bar is the share of the provision's words found in the answer, and the sentence is computed from both.

## 6 What the ontology reveals

Some findings need no query at all. They are properties of the corpus that appear as soon as identity, structure and definitions are explicit.

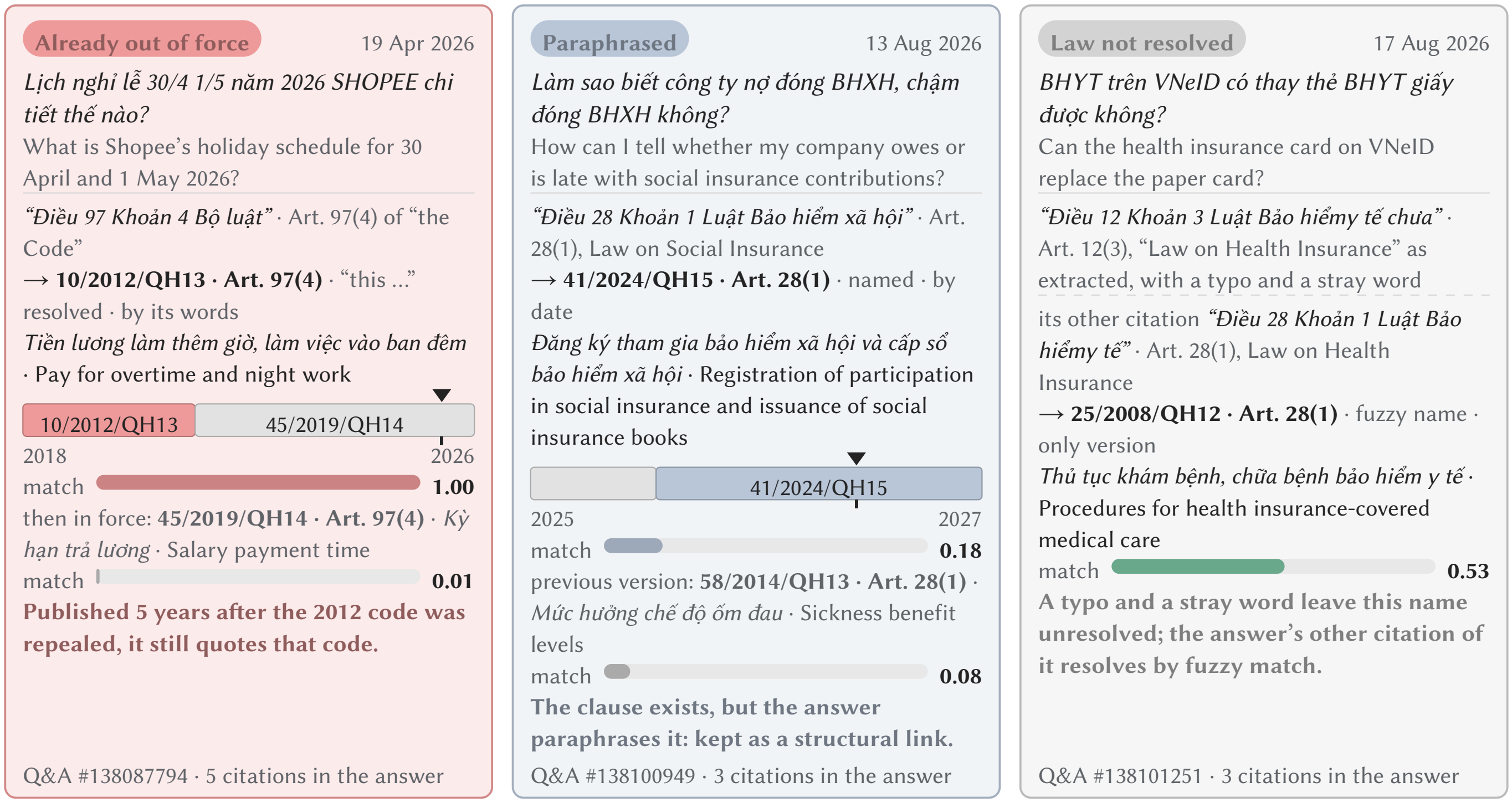

Fig. 8. **Three answers the graph flags.** A citation of a code repealed years before the answer, a paraphrase kept as a structural link, and a law name that cannot be resolved; layout as in Fig. 7.

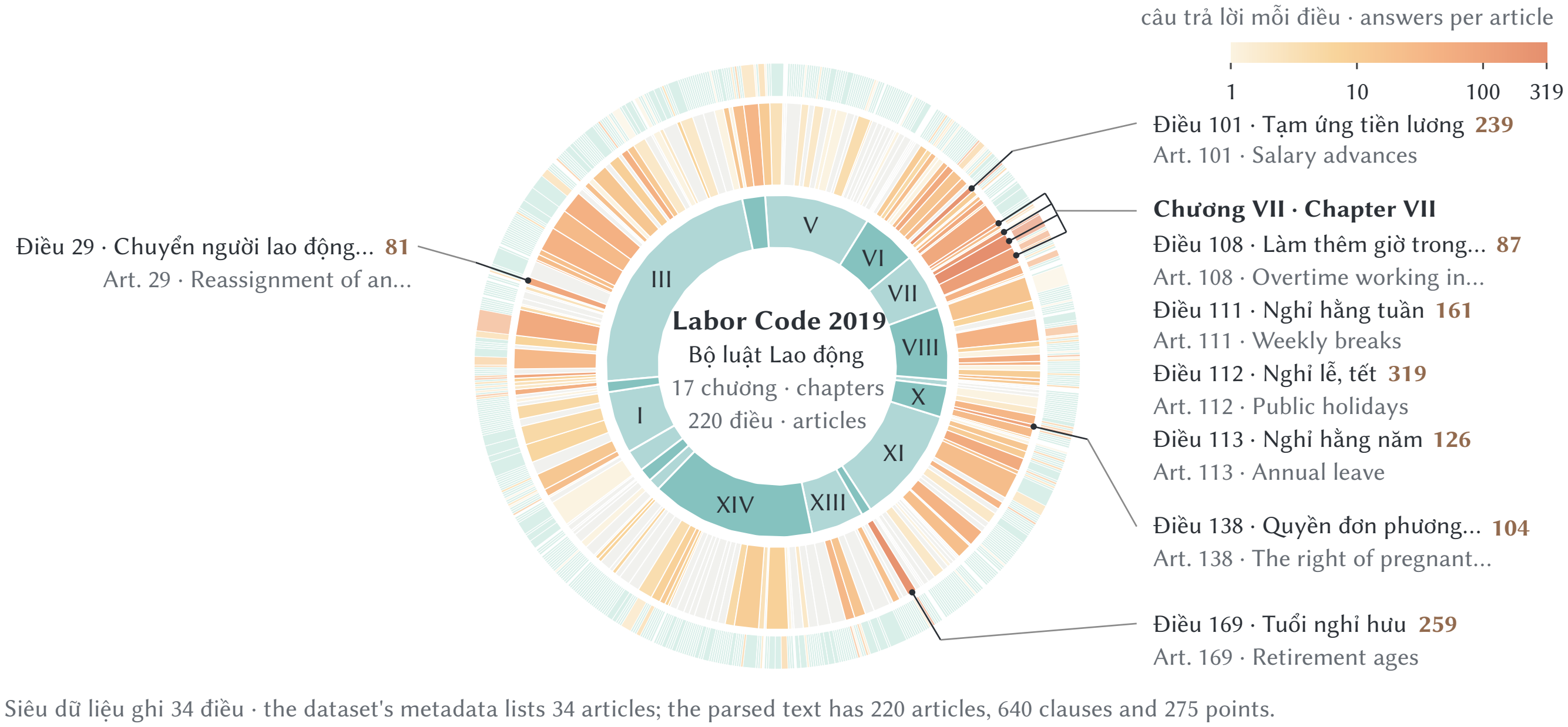

Fig. 9. **Where the questions land.** The 2019 Labor Code as rings, from the inside out: 17 chapters, 220 articles, 640 clauses and 275 points, each sized by its parsed parts. Chapters run clockwise in order; colour shows how many answers cite each article (and each clause). Call-outs name the 8 most-asked articles, and neighbouring articles of one chapter share a call-out. The dataset's metadata lists 34 articles for the same code.

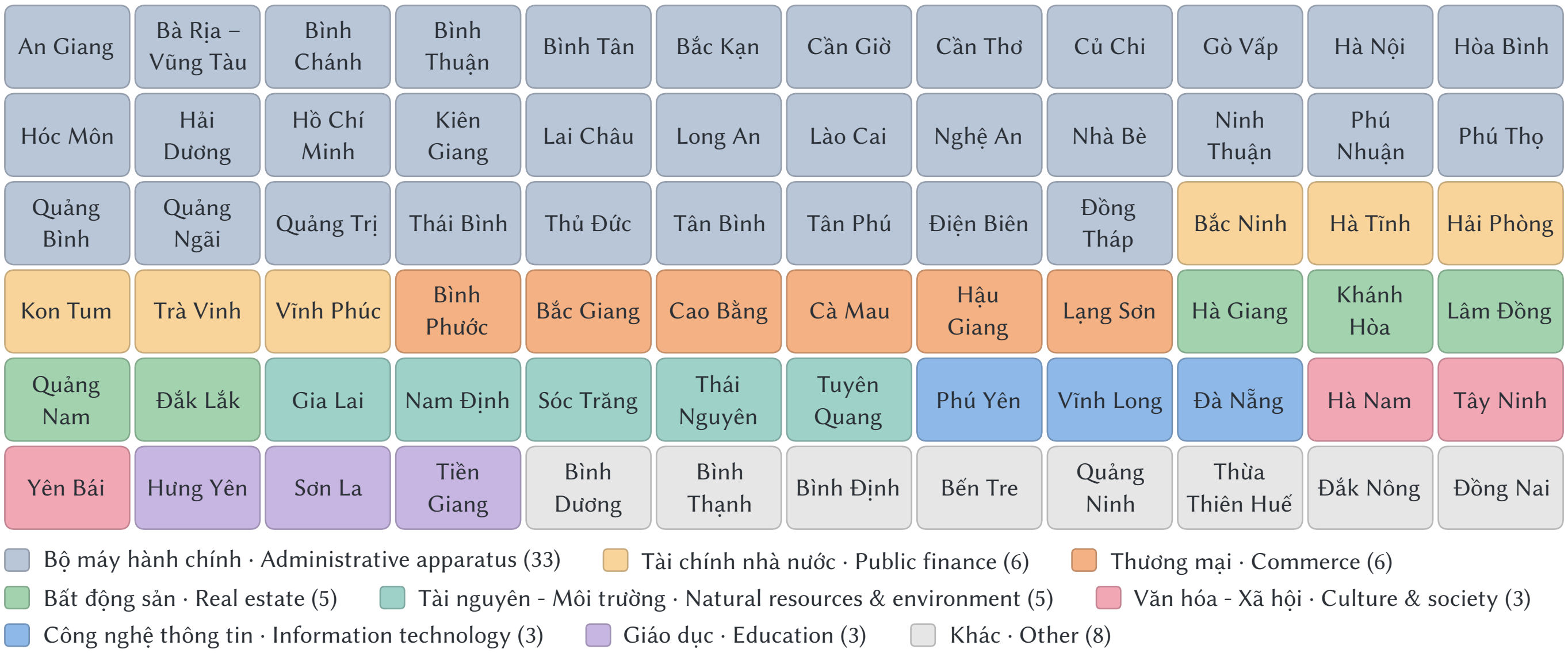

Fig. 10. **One number, 72 decisions.** Every work numbered `01/2024/QĐ-UBND`, one tile per issuing authority (55 provinces, 6 cities and 11 urban or rural districts), coloured by the legal area of the decision. Without its issuer, the number identifies none of them.

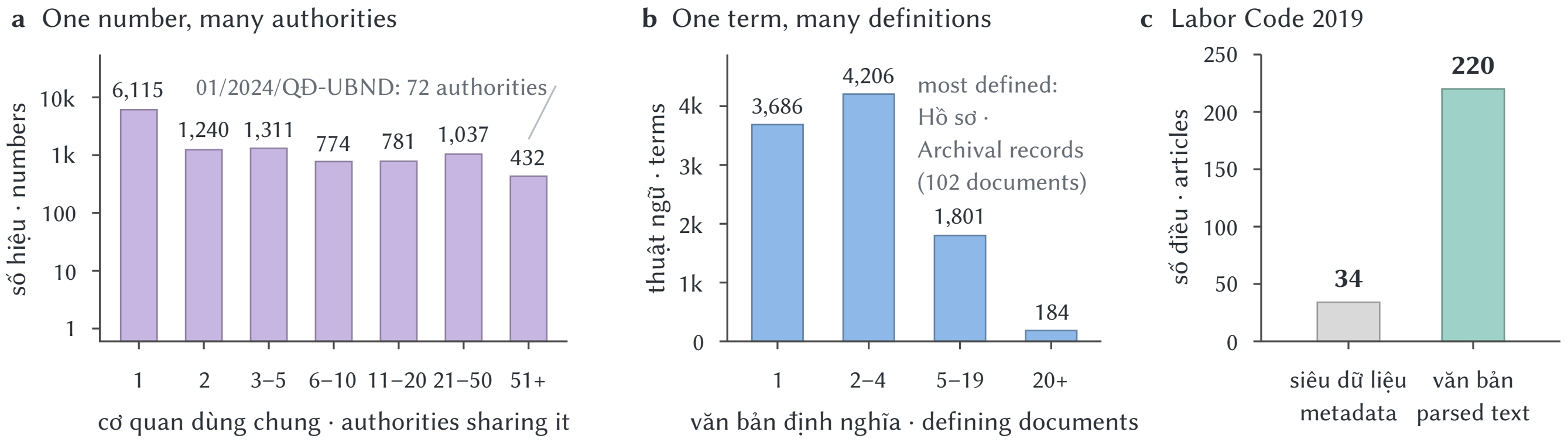

Fig. 11. **Three things a flat table hides.** (a) Of 11,690 full numbers used by local authorities, 5,575 are shared by two or more, one by 72 (log scale). (b) 63% of the 9,877 defined terms have definitions in two or more legal documents. (c) The dataset lists 34 articles for the 2019 Labor Code, whose text has 220.

**Structure outweighs metadata.** For 98,146 of the 319,024 documents with parsed articles (31%), the article count in the metadata is lower than the parsed text shows. The 2019 Labor Code is listed with 34 articles; its text has 220 (Fig. 9). Questions concentrate on a few of them: 31,038 answers cite 13,716 distinct articles across the corpus, yet the most asked, Article 112 of the Labor Code, *Nghỉ lễ, tết* (Public holidays), alone is cited by 319 answers, and the ten most cited account for 4.5% of answer–article links (Table 9).

**Document numbers are not identifiers.** Documents are numbered per issuing authority, so 231,510 local works need their issuer to be unique. Of 11,690 full numbers (number, year and type) used by local authorities, 5,575 (48%) are shared by two or more (Fig. 11 a), and `01/2024/QĐ-UBND` names 72 decisions (Fig. 10).

**Terms are defined many times.** 9,877 terms have at least one validated defining clause, and 6,191 of them (63%) are defined in two or more legal documents (Fig. 11 b). The most redefined, *Hồ sơ (trong hoạt động lưu trữ)* (Archival records), has definitions in 102. What a term means depends on which document an answer applies, and the `defines` edges make that choice explicit.

**The same ontology, twelve laws.** Fig. 12 places the twelve laws with the most text-verified answers side by side, each sampled by the same rule from the same graph; the panel titles give how many answers cite each law with a verified provision. Six of them are versions the dataset records as replaced, such as the 2005 Civil Code and the 2012 Labor Code: answers keep quoting them long after their successors took effect (Section 7.1).

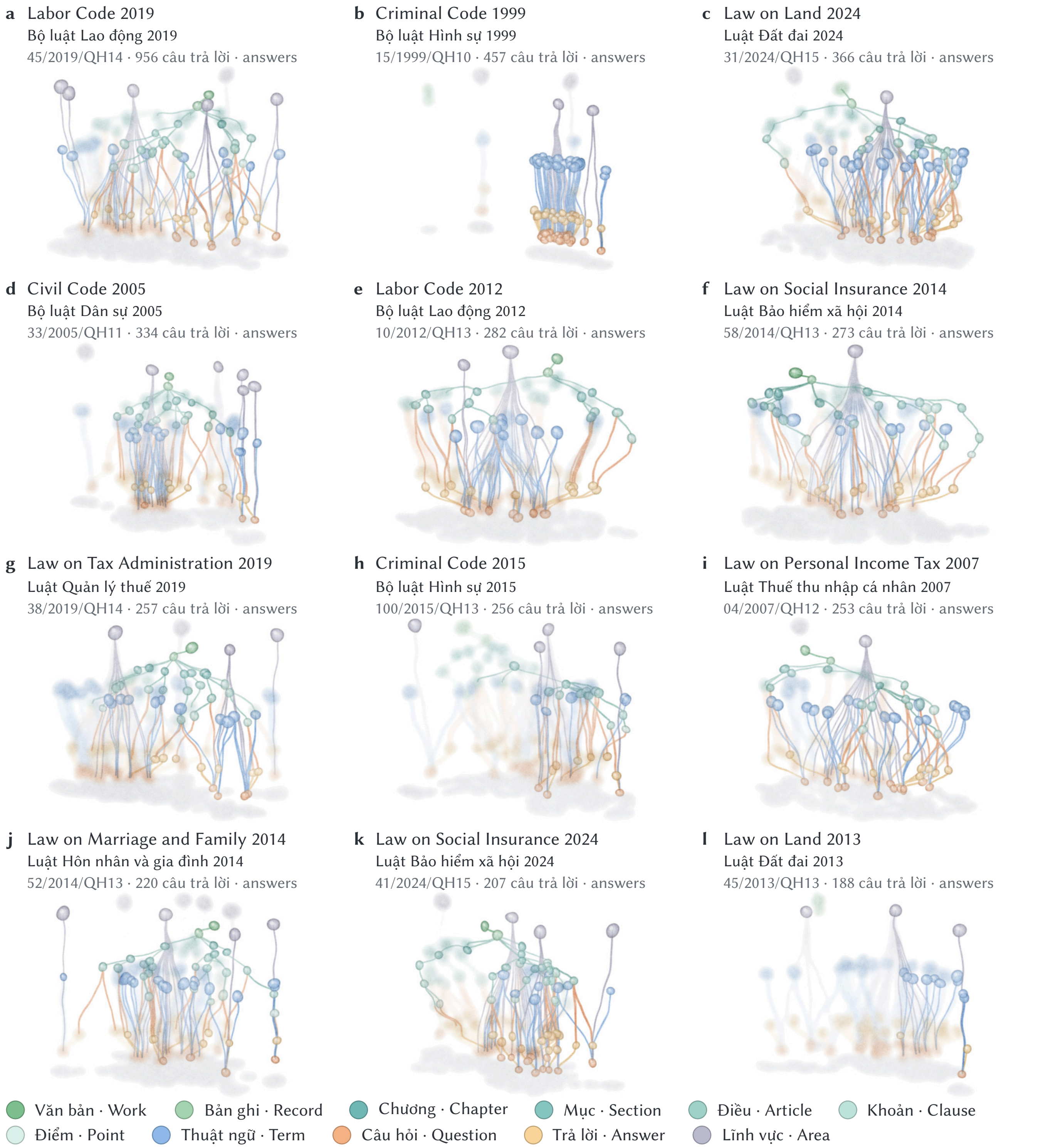

Fig. 12. **Twelve laws as the graph sees them.** The twelve laws with the most text-verified answers, in that order. For each law: its most-cited articles with the clauses and points that answers cite, two verified answers per article with their questions, the terms those questions mention, and their legal areas (87 to 126 nodes per law). Colours mark classes as in Fig. 1; panel titles give the law and the number of answers that cite it with a verified provision.

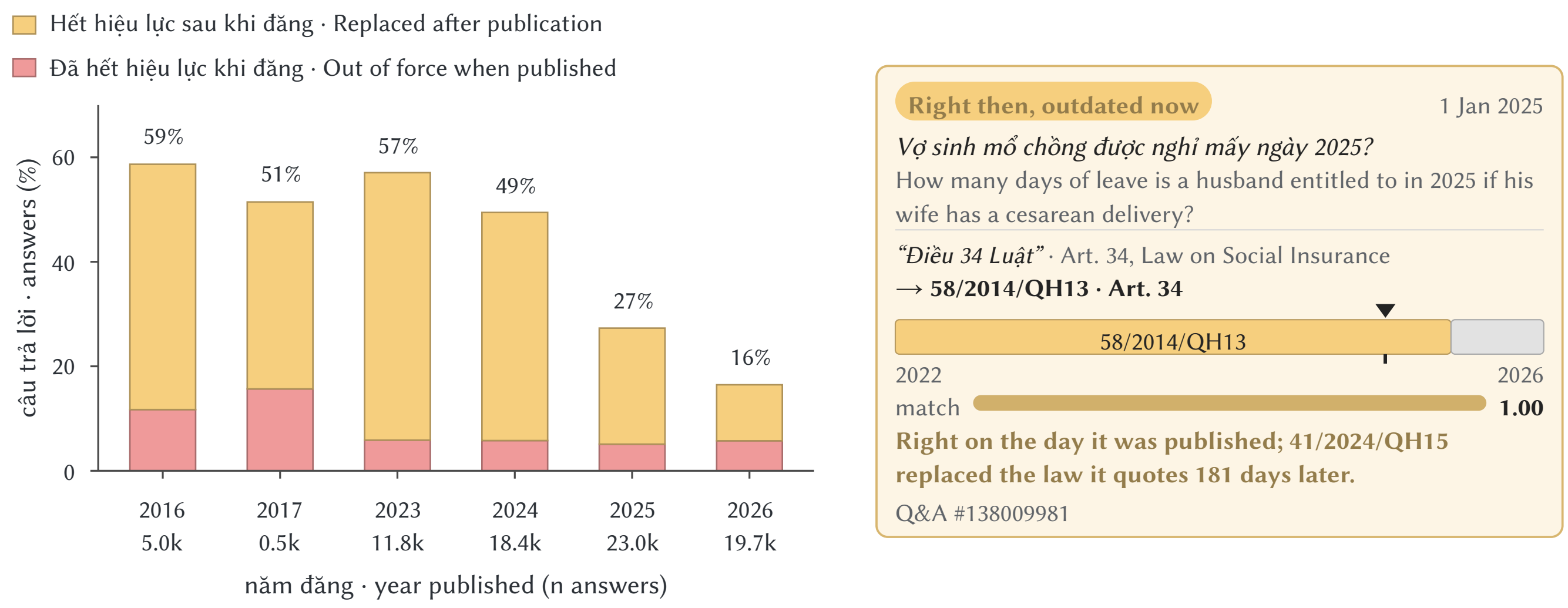

Fig. 13. **A third of the answers now cite a law that is no longer in force.** Left: share of answers citing at least one such law, by year of publication; light, the law was in force at publication and was replaced later; dark, every such law was already out of force at publication. Years with fewer than 100 answers are omitted. Right: one of them, drawn like Fig. 7.

## 7 Findings

With citations grounded in dated versions, three questions about the Q&A become measurable: how answers age, how article numbers change meaning, and which version a writer had in mind.

### 7.1 Answers age

An answer is right about the law on the day it is published. Of the 78,361 answers whose citations resolve, 28,540 (36%) cite at least one law that is no longer in force (Fig. 13). The share is 59% for answers published in 2016 and 57% for 2023, and it falls to 27% in 2025 and 16% in 2026, so even answers less than a year old have aged. Most are not mistakes: 23,803 answers cite a law that was in force when they were written and has been replaced since. Others were out of date from the start: 8,116 cite a law that had already expired or been repealed when they were published, and 4,817 cite a law not yet in force on that date. These groups overlap, because one answer can cite several laws.

Answer #138009981 (1 Jan 2025) asks *"Vợ sinh mổ chồng được nghỉ mấy ngày 2025?"* ("How many days of leave is a husband entitled to in 2025 if his wife has a cesarean delivery?"). It quotes Article 34 of Law 58/2014/QH13, the *Luật bảo hiểm xã hội* (Law on Social Insurance), word for word (match 1.00). The law was in force on that day; Law 41/2024/QH15 replaced it on 01/07/2025. The answer was right when it was written and is out of date now, and only the version graph can tell the two apart. Section 8 turns all 28,540 such answers into a review queue.

Fig. 14 shows the same pattern law by law. In every family most citations name the version in force on the answer's publication date; the share naming a version already replaced or not yet in force ranges from 0.4% (Law on Housing) to 16% (Law on Corporate Income Tax). Across the twelve families, 942 of these 948 citations (99%) are ringed: the answer's own words match the version it cites better than the version in force (Section 7.3). For the Labor Code, 171 of 2,548 citations name a version not then in force.

### 7.2 Same number, different rule

When a law is replaced, its article numbers stay familiar while their content moves. The Labor Code shows how far: of 178 article numbers present in both parsed codes, 155 (87%) share less than a quarter of their word 3-grams (Jaccard similarity below 0.25). Of 186 parsed articles of the 2012 code, 113 have a successor in 2019 with a 3-gram similarity of at least 0.3, and 92 of these changed number. Fig. 15 zooms into rest time. *Nghỉ hằng năm* (Annual leave) was Article 111 in 2012 and is Article 113 in 2019; Article 111 is now *Nghỉ hằng tuần* (Weekly breaks), and Article 112 is now *Nghỉ lễ, tết* (Public holidays). Answers follow the new numbering, with 161, 319 and 126 answers citing 2019 Articles 111, 112 and 113 against 20, 11 and 2 for the same numbers in 2012. A reader who looks up an old answer's "Article 111" in today's code finds a different rule.

The problem is not confined to one code. Of 11,379 citations that resolve to an older version of a multi-version law, 4,357 are confirmed by their words. For 2,525 of these (2,138 answers, 61 law families), the same article number in the latest version either does not exist (1,287) or says something the answer does not quote; Table 7 lists the laws most affected.

### 7.3 The words give the version away

A date is a weak signal for which version a writer meant. Answer #138026329 (4 Jun 2024) asks *"Thời gian nghỉ hè của giáo viên tiểu học là bao lâu?"* ("How long is the summer break for primary school teachers?") and cites *"Điều 111 Bộ luật lao động"* (Art. 111, Labor Code).

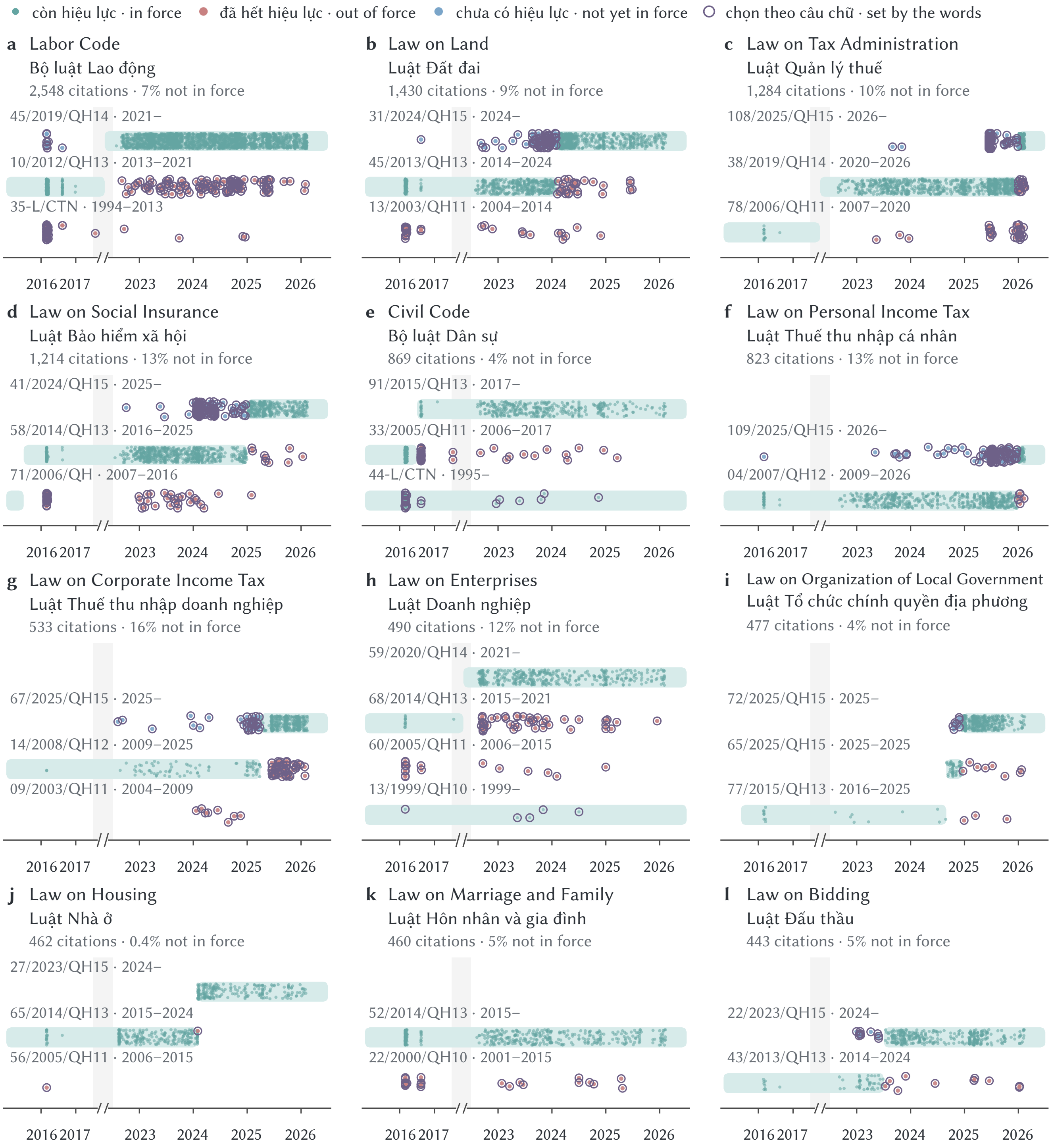

Fig. 14. **Who cites which version, when.** For the twelve law families with the most citations and more than one dated version, each dot is one answer's citation, placed at its publication date in the lane of the version it cites; bands show when each version was in force, as recorded in the dataset. Red dots cite a version that was already out of force, blue ones a version not yet in force, and rings mark citations whose version was chosen by the answer's words. Years 2018–2022, with few answers, are compressed (axis break). The Criminal Code, whose dates in the dataset are unreliable, and the Law on Handling Administrative Violations, with a single dated version, are left out.

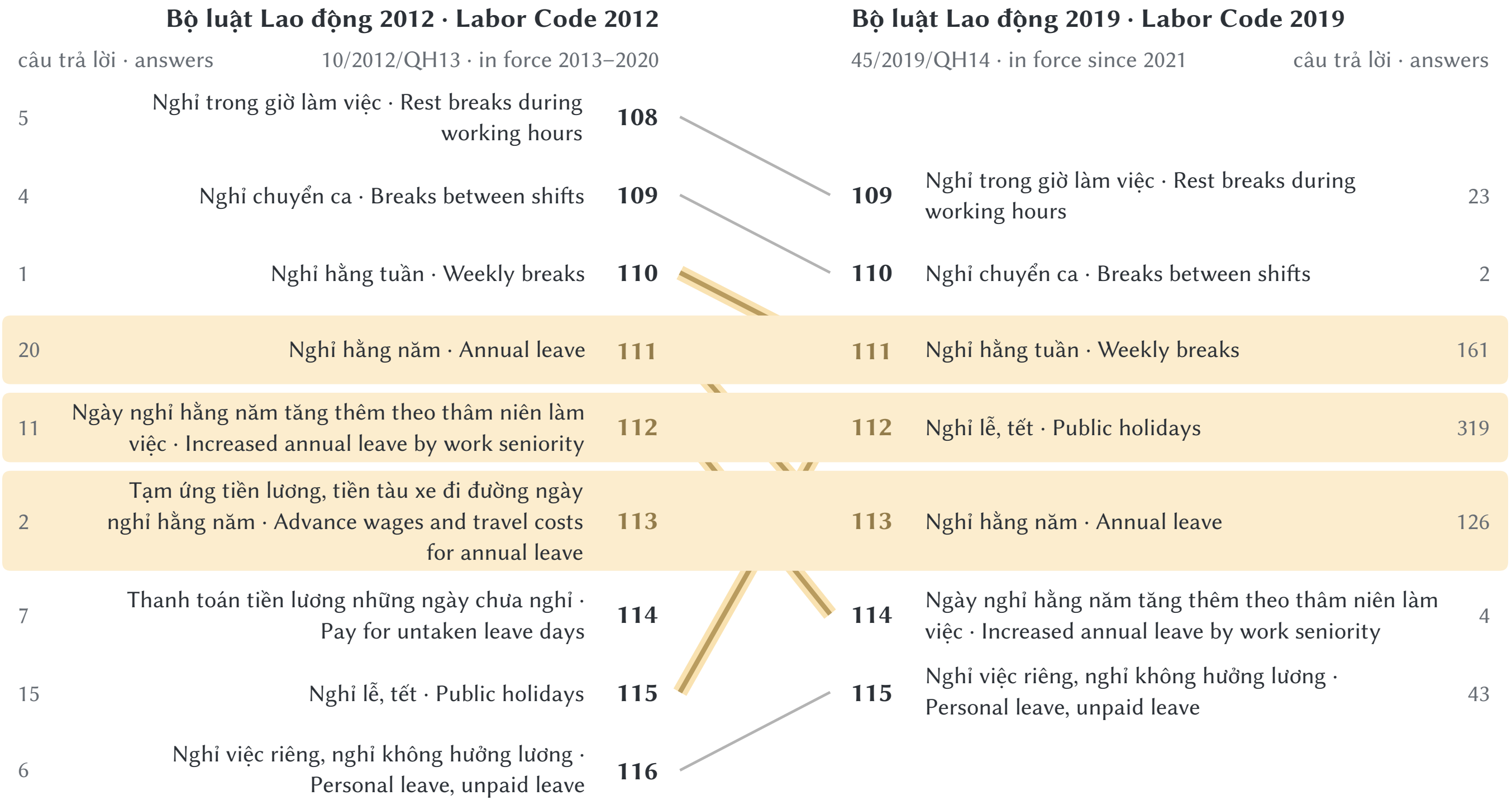

Fig. 15. **Same number, different rule.** Articles of the 2012 Labor Code (left) and the 2019 Labor Code (right) in the rest-time chapter, aligned by number; lines join articles with the same title or similar text. Annual leave moved from Article 111 to Article 113, and Article 111 now governs weekly breaks; the numbers at the outer edges count the answers that cite each article.

Table 2. **Three interfaces, one graph.** Each call returns node identifiers with their provenance, so its result can be traced like Fig. 5.

| Interface | Call | Result |
|---|---|---|
| MCP | `search_terms("người lao động")` | t:630 Người lao động · Employee ⟶ 5 defining clauses, 3,339 questions |
| CLI | `vnkg trace a:138008514` | 43 provenance hops: Q ⟶ A ⟶ Art. 35(1)(d) ⟶ 45/2019/QH14 ⟶ terms |
| API | `GET /api/terms?query=hợp đồng lao động` | t:641 Hợp đồng lao động · Labour contract ⟶ 681 questions |

By date, the citation would resolve to the 2019 code, whose Article 111 is *Nghỉ hằng tuần* (Weekly breaks), and the answer matches that article at only 0.13. Its words match Article 111 of the 2012 code, *Nghỉ hằng năm* (Annual leave), at 1.00. A teacher's summer break is a question about annual leave, and the answer quotes the repealed code's article verbatim.

Text evidence re-versions 2,520 citations in 2,013 answers (Table 8), in both directions.

Among the ten largest flows, 800 citations move to an older version and 668 to a newer one.

The largest moves 517 citations from the 2015 Criminal Code (Law 100/2015/QH13) back to the 1999 code (Law 15/1999/QH10); 442 of them come from answers published in 2016 and 2017, before the 2015 code took effect on 1 January 2018. Closer to home, 116 answers that resolve to the 2019 Labor Code by their date turn out to quote the repealed 2012 code.

## 8 Serving and learning from the graph

A graph that grounds answers is useful only if people and agents can ask it things. Following Auto Ontology, the graph is served through three interfaces: tools over the Model Context Protocol [Anthropic 2024] for agents, a command line for analysts, and a REST API for applications. Table 2 shows one call through each, with results computed on our graph. Each result carries the identifiers and provenance of the nodes it returns, so an agent that cites the graph can be checked in turn.

Learning closes the loop. Every rule that fails leaves a trace that can become a proposal, and nothing enters the graph as validated without evidence. Three queues come out of the current build:

- 453 alias proposals (score ≥ 90) ⟶ +1,651 citations if certified
- 2,520 text-evidence version fixes become resolver rules
- 28,540 answers queued for review: cite law no longer in force

The queue is the practical output for an editor: each item names the answer, the law it cites, the law that replaced it, and the date the change took effect.

## 9 Related work

**Legal documents and identifiers.** Akoma Ntoso models the structure of legislative text [Palmirani and Vitali 2011], and the European Legislation Identifier gives laws, their expressions and their subdivisions stable identifiers [Council of the European Union 2012], building on the work and expression distinction of FRBR [IFLA Study Group on the Functional Requirements for Bibliographic Records 1998]. We adopt both and add what Q&A needs: citations as edges into that structure, each with its evidence.

**Legal ontologies and linked legal data.** Legal ontologies have long organised norms and legal concepts [Breuker et al. 2004], and linked-data projects connect legislation and case law across countries [Filtz et al. 2021]. Our graph applies these ideas to Vietnamese law and connects statutes, Q&A, terms and assistance articles down to the level of points.

**Knowledge graphs and provenance.** We follow common practice for knowledge graphs [Hogan et al. 2021], record provenance in the spirit of PROV-O [Lebo et al. 2013], and model terms and areas with SKOS [Miles and Bechhofer 2009] and Q&A with schema.org [Guha et al. 2016].

**Legal NLP and question answering.** Benchmarks such as LexGLUE [Chalkidis et al. 2022] and surveys of legal AI [Zhong et al. 2020] centre on models, and Vietnamese legal question answering has been studied as retrieval of the relevant articles [Kien et al. 2020]. Retrieval-augmented generation [Lewis et al. 2020] grounds answers in retrieved text but inherits whatever that text cites. The graph supplies the missing layer: which provision, in which version, and whether its words are really there. It is built with DuckDB [Raasveldt and Mühleisen 2019].

## 10 Assumptions and limitations

- **Snapshot.** All counts refer to the dataset snapshot used here. "In force" follows the dataset's status metadata and the repeal sentences we extract, not a check against the official gazette.
- **Assistance data.** The `htpl` records are assistance articles rather than dialogues, and we model them as articles with intents.
- **Auto Ontology.** The five stages and three interfaces follow NVIDIA Auto Ontology, but we did not deploy its service; the calls in Table 2 were evaluated by our own implementation of those interfaces over this graph.
- **Parsing.** The 2012 Labor Code record parses into 186 of its 242 articles (numbers 118–132, 194–234 are missing), so Fig. 15 stays inside articles parsed in both codes.
- **Lexical verification.** Word 5-gram containment confirms quotation, not interpretation: a correct paraphrase stays structural, and a quotation used out of context would pass.
- **Translations.** English for statutes and law names comes from the dataset's English records; English for Q&A titles, for the questions in Fig. 6, Fig. 7 and Fig. 8 and for two 2012 article titles is machine-assisted.
- **Galleries.** The cards in Fig. 7 and Fig. 8 were chosen to show one recent answer per verdict, not sampled at random; how often each verdict occurs is given in Section 7.1 and Section 7.3. The laws in Fig. 12 and Fig. 6 are the twelve with the most text-verified answers; each answer in Fig. 6 is the most recent one for its law that quotes a provision word for word, cites three to eight sources and is not shown elsewhere in the paper.
- **Dates.** Fig. 14 places versions by the effective and expiry dates recorded in the dataset; citations to placeholder versions without a resolved number are left out of the lanes.
- **Missing provisions.** Citations that name a provision absent from the parsed law fall mostly in parse gaps (such as the 2012 Labor Code's), so we count them but do not report them as wrong citations.
- **Issuer names.** Two issuer names that are truncated in the source records are shown in full in Fig. 10: Hải Phòng and Bà Rịa – Vũng Tàu.

## 11 Conclusion

A legal answer is a claim about the law at a moment in time. Making the claim explicit (which provision, in which version, and whether its words are there) turns a corpus of strings into a graph that can say when an answer was right, when it stopped being right, and when an article number has come to mean something else. The same structure that grounds one answer about notice periods exposes 28,540 answers that have aged, 2,525 citations whose article number now holds a different rule, and 2,520 citations whose words point to another version than their date. Every figure and number in this paper is regenerated from the graph by the code in the source bundle (Section A).

## A Reproducibility

The source bundle rebuilds everything in this paper. The pipeline `vnkg/` implements the five stages: `discover.py`, `ground.py`, `core.py` (validation and the DuckDB store), `findings.py`, `expose.py` and `cli.py` (MCP, CLI and REST), `learn.py`, and `sample.py` for the subgraph of Fig. 1; `run.py` runs them all. `assemble_facts.py`, `paper_facts.py` and `facts_display.py` write `facts.json` and `data/`, from which the paper reads every number. Four `gallery_data*.py` scripts extract the gallery examples from the graph; eleven `fig_*.py` scripts draw the figures with Plotly through `plotlib.py` (palette in `figlib.py`), with node positions in Fig. 1 from force-directed placement [Fruchterman and Reingold 1991], and `paper.sh` compiles `main.typ` with Typst and the faithful-acmart template. The graph builds in about 4.3 minutes on one NVIDIA GB300 node; the figures and the paper take a few minutes more.

## B Ontology reference

Table 3 and Table 4 give every class and relation with the vocabulary it aligns with and its size in the built graph.

## C Reconciliation rules

Table 5 gives one worked mapping per kind of link, and Table 6 lists the rules applied in order, with their outcome on the full corpus.

## D Findings in detail

Table 7 lists the laws whose older versions are cited most often for an article that is missing or different in the latest version (Section 7.2), Table 8 the ten largest re-versionings by text evidence (Section 7.3), and Table 9 the most-asked articles (Section 6).

## E Index of examples

Table 10 lists every answer shown in the paper, in order of appearance, with the provision the graph resolves for the citation shown; `vnkg trace a:`*number* prints the provenance chain behind any row.

Table 3. Classes of the ontology. Chapter and Section are separate classes and share a row.

| Class · Vietnamese | Identity in the data | Aligns with | Nodes |
|---|---|---|---|
| LegalWork · *Văn bản (tác phẩm)* | reconciled number key (header › title › field) | eli:LegalResource · FRBR work | 550,654 |
| LegalDocument · *Bản ghi văn bản* | one vbpl record (VN, EN or consolidated) | eli:LegalExpression | 573,851 |
| Chapter, Section · *Chương, Mục* | headings parsed from the text | akn:chapter · akn:section | 253,621 + 40,372 |
| Article · *Điều* | "Điều N." headings | eli:LegalResourceSubdivision · akn:article | 2,213,352 |
| Clause · *Khoản* | numbered paragraphs "1." | akn:paragraph | 4,583,894 |
| Point · *Điểm* | lettered items "a)", "đ)" | akn:point | 3,073,777 |
| LegalTerm · *Thuật ngữ* | tnpl term_id | skos:Concept | 16,701 |
| LegalArea · *Lĩnh vực* | two schemes: VB (vbpl/hdpl) and TN (tnpl) | skos:Concept · ConceptScheme | 79 |
| Agency · *Cơ quan ban hành* | issuing_body or document header | org:Organization | 824 |
| Question · *Câu hỏi* | hdpl id | schema:Question | 109,543 |
| Answer · *Câu trả lời* | hdpl id | schema:Answer | 109,543 |
| AssistanceArticle · *Bài hỗ trợ* | htpl id | schema:Article | 681 |
| AssistanceIntent · *Ý định hỗ trợ* | htpl category + title cues | skos:Concept (intent scheme) | 9 |

Table 4. Relations of the ontology. Every edge also carries its method, status and source record.

| Relation · Vietnamese | Domain ⟶ range | Aligns with | Edges |
|---|---|---|---|
| `contains` · *chứa* | Document ⟶ Chapter / Section ⟶ Article ⟶ Clause ⟶ Point | eli:has_part · Akoma Ntoso hierarchy | 10,165,016 |
| `defines` · *định nghĩa* | Clause ⟶ LegalTerm (scored) | skos:definition (reified) | 47,691 |
| `cites` · *viện dẫn* | Answer / Document / unit / AssistanceArticle ⟶ Work / Article / Clause / Point | eli:cites | 3,727,394 |
| `amends` · *sửa đổi, bãi bỏ* | Document ⟶ Work (amend · repeal · replace · *_part) | eli:amends · eli:repeals | 41,636 |
| `answers` · *trả lời* | Answer ⟶ Question | schema:acceptedAnswer (inverse) | 109,543 |
| `applies_to` · *áp dụng cho* | Document / Question / Term / AssistanceArticle ⟶ LegalArea Intent ⟶ Term / Work / unit | eli:is_about · dcterms:subject | 707,129 |
| `realizes` · *hiện thực hoá* | LegalDocument ⟶ LegalWork | eli:realizes (expression ⟶ work) | 573,851 |
| `mentions` · *nhắc đến* | Question / AssistanceArticle ⟶ LegalTerm | schema:mentions | 286,643 |
| `issued_by` · *ban hành bởi* | LegalDocument ⟶ Agency | eli:passed_by | 280,635 |
| `related_to` · *liên quan* | LegalTerm ⟷ LegalTerm | skos:related | 17,715 |
| `expresses` · *thể hiện* | AssistanceArticle ⟶ AssistanceIntent | task intent (local) | 746 |
| `maps_to` · *ánh xạ* | LegalArea ⟶ LegalArea | skos:exact / close / narrow / relatedMatch | 35 |

Table 5. One worked mapping per kind of link.

| Mapping | Rule | Example | Result |
|---|---|---|---|
| Number ⟷ work | header › title › number field | "Bộ luật số: 45/2019/QH14" · "Law 45 2019 QH14 Labor Code" | 2 records ⟶ wk:45/2019/QH14 |
| Name + date ⟶ version | family member in force on the Q&A date | "Bộ luật Lao động" (Labor Code) in 2016 ⟶ 10/2012/QH13; in 2024 ⟶ 45/2019/QH14 | 25,551 citations |
| Anaphora | nearest preceding named law of the same type | "khoản 3 Điều 101 của Bộ luật này" · "Clause 3, Article 101 of this Code" | 24,651 citations |
| Provincial number | number + issuer from header | 01/2024/QĐ-UBND · 72 authorities | 231,510 works |
| Legal areas | VB scheme ⟷ TN scheme, SKOS relations | "Lao động – Tiền lương" ⟷ "Lao dong Tien luong" · Labour – Wages | 22 exact · 7 close |
| Category meaning | hdpl / vbpl category = area; htpl category = genre ⟶ intent | "tranh-chap" (disputes) ⟶ Dispute · "chinh-sach-moi" (new policy) ⟶ What's new | 681 articles ⟶ 9 intents |
| Term ⟷ definition | tnpl term ⟷ definitional clause ("X là ..." = "X is ...") | "Người lao động" · Employee ⟷ 45/2019/QH14 Art. 3(1) | 35,400 validated · 12,291 candidate · 6,150 homonym |

Table 6. Reconciliation and validation rules, in the order they are applied.

| Step · Vietnamese | Rule | Outcome |
|---|---|---|
| Document identity · *Định danh văn bản* | Number from the text header ("Số: …" = "No.: …") › title › number field; diacritics folded; undated short numbers stay unique; provincial numbers qualified by issuer (metadata or the "ỦY BAN NHÂN DÂN TỈNH …" = "People's Committee of … Province" header); a record whose type contradicts its number's code is split off. | 573,851 records ⟶ 550,654 works · 20,096 merge ≥ 2 records · 231,510 issuer-qualified · 862 split |
| Versions & time · *Phiên bản và hiệu lực* | Families by (type, name) ordered by effective date; whole repeals read from "… hết hiệu lực / bãi bỏ / thay thế" ("expires / is repealed / is replaced") sentences about "… này" ("this …"); partial ones (Điều, khoản, Mẫu, Phụ lục = article, clause, form, appendix) kept apart; a repeal is rejected if the repealer ranks lower, carries another law's name, or contradicts metadata. | 336 families · 102 multi-version · 30,618 whole repeals kept · 1,061 rejected |
| Named citations · *Trích dẫn có tên* | law_type + law_name ⟶ family: exact name › unique prefix › fuzzy match ≥ 92 (rapidfuzz WRatio). | exact 14,528 · prefix 1,864 · fuzzy 169 · unresolved 2,658 |
| Unnamed citations · *Trích dẫn không tên* | Context cascade inside the answer: number next to "Điều N" (Article N) › law name in the sentence › "… này" ("this …") anaphora › preceding reference of the same type › sole reference › the question text. | 44,926 / 48,649 recovered · anaphora 24,651 · number 14,184 · name 2,693 · sole 2,828 |
| Version choice · *Chọn phiên bản* | Explicit number › explicit year › the version in force on the Q&A publication date › the only version; text override when another version matches the answer ≥ 0.5 and ≥ 0.25 better. | number 27,032 · date 25,551 · single 6,185 · year 199 · text override 2,520 |
| Provision lookup · *Tra điều khoản* | Điều / Khoản / Điểm (article / clause / point) path looked up in the parsed hierarchy of the chosen version (block-aware for attached regulations). | exists 49,846 · article only 1,848 · missing 5,683 · no structure 4,110 |
| Text verification · *Kiểm chứng câu chữ* | Word 5-gram containment of the provision text inside the answer; ≥ 0.5 ⟶ validated; lower ⟶ kept as structural and queued for review. | verified 19,012 · statuses: validated 141,630 · structural 32,682 · unresolved 11,730 · flagged 9,793 |

Table 7. Laws whose older versions are cited most often for an article that is missing, or says something else, in the latest version.

| Older version | Law · Vietnamese | Latest version | Citations |
|---|---|---|---|
| 33/2005/QH11 | Civil Code · *Bộ luật Dân sự* | 91/2015/QH13 | 428 |
| 15/1999/QH10 | Criminal Code · *Bộ luật Hình sự* | 100/2015/QH13 | 236 |
| 38/2019/QH14 | Law on Tax Administration · *Luật Quản lý thuế* | 108/2025/QH15 | 224 |
| 58/2014/QH13 | Law on Social Insurance · *Luật Bảo hiểm xã hội* | 41/2024/QH15 | 189 |
| 10/2012/QH13 | Labor Code · *Bộ luật Lao động* | 45/2019/QH14 | 159 |
| 45/2013/QH13 | Law on Land · *Luật Đất đai* | 31/2024/QH15 | 158 |
| 65/2014/QH13 | Law on Housing · *Luật Nhà ở* | 27/2023/QH15 | 122 |
| 24/2004/QH11 | Code of Civil Procedure · *Bộ luật Tố tụng dân sự* | 92/2015/QH13 | 108 |

Table 8. The ten largest re-versionings by text evidence. Each citation moves to the version of the same law whose article its words match.

| Chosen by date or name | Matched by words | Law · Vietnamese | Citations |
|---|---|---|---|
| 100/2015/QH13 | 15/1999/QH10 | Criminal Code · *Bộ luật Hình sự* | 517 |
| 04/2007/QH12 | 109/2025/QH15 | Law on Personal Income Tax · *Luật Thuế thu nhập cá nhân* | 168 |
| 38/2019/QH14 | 108/2025/QH15 | Law on Tax Administration · *Luật Quản lý thuế* | 154 |
| 45/2019/QH14 | 10/2012/QH13 | Labor Code · *Bộ luật Lao động* | 121 |
| 58/2014/QH13 | 41/2024/QH15 | Law on Social Insurance · *Luật Bảo hiểm xã hội* | 116 |
| 92/2015/QH13 | 24/2004/QH11 | Code of Civil Procedure · *Bộ luật Tố tụng dân sự* | 108 |
| 45/2013/QH13 | 31/2024/QH15 | Law on Land · *Luật Đất đai* | 94 |
| 58/2010/QH12 | 129/2025/QH15 | Law on Public Employees · *Luật Viên chức* | 79 |
| 53/2014/QH13 | 46/2024/QH15 | Law on Notarization · *Luật Công chứng* | 57 |
| 67/2025/QH15 | 14/2008/QH12 | Law on Corporate Income Tax · *Luật Thuế thu nhập doanh nghiệp* | 54 |

Table 9. The ten most-asked articles, counted by distinct answers citing them.

| Law | Article | Title · Vietnamese | Answers |
|---|---|---|---|
| 45/2019/QH14 | 112 | Public holidays · *Nghỉ lễ, tết* | 319 |
| 04/2007/QH12 | 3 | Taxable income · *Thu nhập chịu thuế* | 287 |
| 58/2014/QH13 | 2 | Subjects of application · *Đối tượng áp dụng* | 276 |
| 45/2019/QH14 | 169 | Retirement ages · *Tuổi nghỉ hưu* | 259 |
| 45/2019/QH14 | 101 | Salary advances · *Tạm ứng tiền lương* | 239 |
| 12/2022/NĐ-CP | 46 | Violations committed by Vietnamese guest workers and some other related entities · *Vi phạm của người lao động Việt Nam đi làm việc ở nước ngoài và một số đối tượng liên quan khác* | 194 |
| 04/2007/QH12 | 4 | Tax-exempt income · *Thu nhập được miễn thuế* | 179 |
| 41/2024/QH15 | 2 | Compulsory and voluntary social insurance participants · *Đối tượng tham gia bảo hiểm xã hội bắt buộc và bảo hiểm xã hội tự nguyện* | 173 |
| 123/2020/NĐ-CP | 10 | Content of invoices · *Nội dung của hóa đơn* | 163 |
| 45/2019/QH14 | 111 | Weekly breaks · *Nghỉ hằng tuần* | 161 |

Table 10. Every answer shown in the paper, in order of appearance: its number and publication date, and the law version and provision the graph resolves for the citation shown.

| Q&A | Published | Law | Provision | Shown in | Q&A | Published | Law | Provision | Shown in |
|---|---|---|---|---|---|---|---|---|---|
| #138008514 | 14 Mar 2024 | 45/2019/QH14 | Art. 35(1)(d) | Fig. 1, 5 | #138040994 | 22 Dec 2024 | 45/2013/QH13 | Art. 7 | Fig. 6l |
| #138017201 | 10 Jan 2024 | 45/2019/QH14 | Art. 101(3) | Sec. 4 | #138100449 | 13 Aug 2026 | 41/2024/QH15 | Art. 25(1) | Fig. 7 |
| #138066888 | 11 Aug 2026 | 45/2019/QH14 | Art. 112(1) | Fig. 6a | #138099228 | 5 Aug 2026 | 45/2019/QH14 | Art. 113(1) | Fig. 7 |
| #6969 | 13 Feb 2017 | 15/1999/QH10 | Art. 157 | Fig. 6b | #138099436 | 1 Aug 2026 | 137/2020/NĐ-CP | Art. 11 | Fig. 7 |
| #138076129 | 2 Jan 2026 | 31/2024/QH15 | Art. 75(5) | Fig. 6c | #138099545 | 10 Aug 2026 | 04/2007/QH12 | Art. 3(2) | Fig. 7 |
| #7068 | 13 Feb 2017 | 33/2005/QH11 | Art. 25 | Fig. 6d | #138095709 | 29 Jun 2026 | 108/2025/QH15 | Art. 26 | Fig. 7 |
| #138035189 | 1 Oct 2024 | 10/2012/QH13 | Art. 64 | Fig. 6e | #138095360 | 23 Jun 2026 | 38/2019/QH14 | Art. 124 | Fig. 7 |
| #138071343 | 27 Nov 2025 | 58/2014/QH13 | Art. 43 | Fig. 6f | #138087794 | 19 Apr 2026 | 10/2012/QH13 | Art. 97(4) | Fig. 8 |
| #138087381 | 16 Apr 2026 | 38/2019/QH14 | Art. 44 | Fig. 6g | #138100949 | 13 Aug 2026 | 41/2024/QH15 | Art. 28(1) | Fig. 8 |
| #138064101 | 26 Sep 2025 | 100/2015/QH13 | Art. 177 | Fig. 6h | #138101251 | 17 Aug 2026 | unresolved | Art. 12(3) | Fig. 8 |
| #138079788 | 2 Feb 2026 | 04/2007/QH12 | Art. 3(1) | Fig. 6i | #138009981 | 1 Jan 2025 | 58/2014/QH13 | Art. 34 | Fig. 13 |
| #138033828 | 20 Sep 2024 | 52/2014/QH13 | Art. 34 | Fig. 6j | #138026329 | 4 Jun 2024 | 10/2012/QH13 | Art. 111 | Sec. 7.3 |
| #138098515 | 21 Jul 2026 | 41/2024/QH15 | Art. 47 | Fig. 6k | | | | | |